\documentclass[10pt,twocolumn]{article} 
\usepackage{simpleConference}
\usepackage{cite}
\usepackage{amsmath,amssymb,amsfonts}
\usepackage{algorithmic}
\usepackage{graphicx}
\usepackage{hyperref}
\usepackage{textcomp}
\usepackage{xcolor}
\def\BibTeX{{\rm B\kern-.05em{\sc i\kern-.025em b}\kern-.08em
		T\kern-.1667em\lower.7ex\hbox{E}\kern-.125emX}}

\usepackage{caption}
\usepackage{listings}
\usepackage{algorithm}

\usepackage{subcaption}
\usepackage{color}

\usepackage{listings}
\usepackage{authblk}
\usepackage{amsmath}
\newcommand{\norm}[1]{\left\lVert#1\right\rVert}

\definecolor{deepblue}{rgb}{0,0,0.5}
\definecolor{deepred}{rgb}{0.6,0,0}
\definecolor{deepgreen}{rgb}{0,0.5,0}

\definecolor{codegreen}{rgb}{0,0.4,0}
\definecolor{codegray}{rgb}{0.5,0.5,0.5}
\definecolor{codepurple}{rgb}{0,0,0.5}
\definecolor{backcolour}{rgb}{0.95,0.95,0.92}

\DeclareFixedFont{\ttm}{T1}{txtt}{m}{n}{8}

\lstdefinestyle{mystyle}{
	backgroundcolor=\color{white},
	commentstyle=\color{codegreen},
	keywordstyle=\color{deepred},
	numberstyle=\tiny\color{codegray},
	stringstyle=\color{codepurple},
	basicstyle=\ttfamily\small,
	breakatwhitespace=false,         
	breaklines=true,                 
	captionpos=b,                    
	keepspaces=true,                 
	numbers=left,                    
	numbersep=5pt,                  
	showspaces=false,                
	showstringspaces=false,
	showtabs=false,                  
	tabsize=2
}

\begin{document}

\title{GradVis: Visualization and Second Order Analysis of Optimization Surfaces during the Training of Deep Neural Networks}

\author[1,3]{Avraam Chatzimichailidis}
\author[1,2]{Franz-Josef Pfreundt}
\author[3]{Nicolas R. Gauger}
\author[1,4]{Janis Keuper}

\affil[1]{Competence Center High Performance Computing, Fraunhofer ITWM, Kaiserslautern, Germany}
\affil[2]{Fraunhofer Center Machine Learning, Germany}
\affil[3]{Chair for Scientific Computing, TU Kaiserslautern, Germany}
\affil[4]{Institute for Machine Learning and Analytics, Offenburg University, Germany}

\renewcommand\Authands{ and }
\date{}
\maketitle
\thispagestyle{empty}

\begin{abstract}
Current training methods for deep neural networks boil down to very high dimensional and
non-convex optimization problems which are usually solved by a wide range of
stochastic gradient descent methods. While these approaches tend to work 
in practice, there are still many gaps in the theoretical understanding of key aspects like 
convergence and generalization guarantees, which are induced by the properties of the optimization surface (loss landscape).
In order to gain deeper insights, a number of recent publications proposed methods to visualize and analyze the optimization surfaces. 
However, the computational cost of these methods are very high, making it hardly possible to use them on larger networks.\\    
In this paper, we present the GradVis Toolbox, an open source library for efficient and scalable visualization and analysis of deep neural network
loss landscapes in Tensorflow and PyTorch. Introducing more efficient mathematical formulations and a novel parallelization scheme, GradVis
allows to plot 2d and 3d projections of optimization surfaces and trajectories, as well as high resolution second order gradient information for large networks. 
\end{abstract}

\section{Introduction}
Training neural networks is a NP-hard problem \cite{Blum1988}, as it requires
finding minima of a high-dimensional non-convex loss function.  In practice,
most theoretical implications of the non-convexity are simply ignored by the deep 
learning community and it has become the standard approach to use
methods that only provide convergence guarantees for convex problems. In most cases, 
optimizers such as Stochastic Gradient Descent (SGD) \cite{kiefer1952} are actually able
to converge into local minima. However, while this approach appears to be working in practice,
there is still a wide range of open theoretical questions associated with these optimization
problems. Some of which could have major impact on the practical application, like the
current discussion about the affect of minima shapes on the generalization abilities of the 
resulting model.\\
\textbf{The wide minimum hypothesis:} 
It has been
widely believed that geometrically wide minima in the loss surface would lead to better generalizing models, as argued by
\cite{DBLP:conf/colt/HintonC93},\cite{DBLP:journals/corr/ChaudhariCSL16} and
\cite{DBLP:journals/corr/KeskarMNST16}.
On the contrary, \cite{jastrzkebski2018relation} showed that sharp minima can
generalize just as well.
\begin{figure}[h!]
	\centering
	\includegraphics[width=0.76\linewidth]{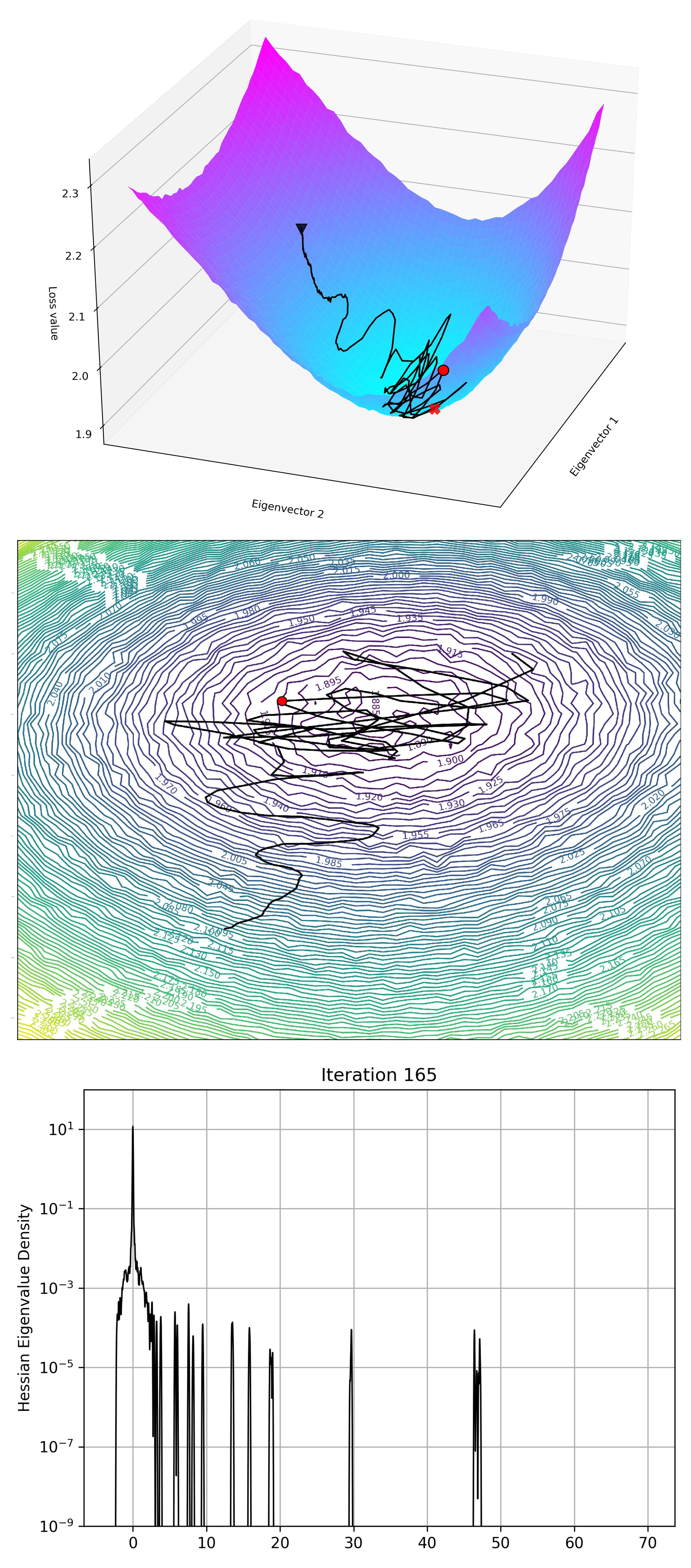}
	\caption{Using GradVis to plot the 3D and 2D Loss landscape, full gradient trajectory and eigenvalue density of the Hessian for a single iteration. Example shows a LeNet training on CIFAR10. Video of this visualization over all 
		ittertions is available at: \href{https://youtu.be/0AKSjp-SHlo}{https://youtu.be/0AKSjp-SHlo}  }
	\label{fig:iter3}
\end{figure}
These seemingly contradicting results can be explained
by their different approaches to measure flatness. 
\cite{DBLP:journals/corr/DinhPBB17} showed that one can
reparametrize the loss function, thereby altering the geometry of the parameter
space without affecting how the network evaluates unseen data. Hence,
one can simply reparametrize the weights in a neural network layer
without changing the value of the loss function\footnote{ This is true for positively
	homogeneous layer activations (e.g. ReLUs) and batch-norm layers
	\cite{DBLP:journals/corr/IoffeS15}.  As a result, one is able to make minima
	seem arbitrarily sharp or wide.}{.}\\
This current discussion shows, that it is necessary to gain further theoretical insights into these optimization problems. A crucial contribution to such efforts 
could come from tools for the visualization and high resolution quantification of loss surface properties during the training process.

\subsection{Related Work I: Loss surface visualization}
First attempts to visualize loss landscapes were presented by
\cite{goodfellow2014qualitatively}. The authors drew samples along a line in weight space,
connecting the initialization point to the converged minimum. Along this line,
the loss is calculated and plotted. As one result of their experiments, they were able to 
show that the loss along this line
follows a monotonically decreasing path. 
\cite{DBLP:journals/corr/abs-1712-09913} introduced a new method on how to
visualize the loss landscape. Their main contribution is to use filter-wise
normalization to combat the scaling invariance of the weights. The authors also use
a PCA \cite{PCA} on the training paths, reducing the problem onto a 2D plane. 
The landscape is then plotted by spanning a plane in
parameter space around the converged point $\theta^*$ and taking the loss
$L(\theta)$ at each point on this plane. Since PCA's complexity is cubic, both in the number of samples and their dimensionality, and comes with high memory demands, it is quite difficult to implement this approach for larger networks.
Additionally, this approach has been 
criticised by \cite{antognini2018pca}, who
showed that  performing PCA on high-dimensional random walks always results in
Lissajous trajectories. They argue that the training trajectory, if plotted
using PCA, always performs the same patterns, thereby rendering the information
one can extract from the trajectory useless.

\subsection{Related Work II: Second order properties of the loss surface}
Another approach to gain further insight into the loss landscape is to compute the eigenvalues of the Hessian
at different points in parameter space. The eigenvalues of the
Hessian are a measure of curvature of the loss landscape. Positive eigenvalues
correspond to positive curvature along its corresponding eigenvector, meaning
the network is inside a minimum in this direction. Also its magnitude
corresponds to how steep this minimum is.  Calculating the eigenvalues of the
Hessian of a neural network has a complexity of $O(N^3)$. For regular networks
with parameters in the range of $10^6-10^8$, this is infeasible to compute. Especially for all of the usual $10^5$ evaluations points during the iterations of the  optimization process.
Therefore, in order to compute the full eigenvalue spectrum, we adopt two common compute "tricks" from other fields: 
First we are able to compute the Hessian-vector product efficiently in $O(N)$
instead of $O(N^2)$ without having to store the Hessian during this process.
This is done by using the R-operator \cite{Pearlmutter94fastexact}, which is a
linear operator that wraps around the forward and backward propagation in order
to compute the Hessian-vector product.  Second, by using the Lanczos algorithm \cite{lanczos},
it only requires the Hessian-vector product to compute the eigenvalues. As
\cite{quadrature} shows, one can approximate the full eigenvalue spectrum by
only performing $m$ Lanczos iteration steps and by then diagonalizing the
resulting $m\times m$ tridiagonal matrix. This procedure is done $k$ times with different
starting vectors. Using the resulting eigenvalues and eigenvectors of this
tridiagonal matrix, one can approximate the full eigenvalue spectrum using
Gaussians to high accuracy in only $O(Nmk)$. The first paper to
apply this method for neural networks was \cite{firstquad}. Through out this paper,
we largely follow the notation of \cite{secondquad}, using the stochastic Lanczos
quadrature algorithm.

\subsection{Contribution}
In this paper, we present an open source toolbox\footnote{available at: \href{https://github.com/cc-hpc-itwm/GradVis}{https://github.com/cc-hpc-itwm/GradVis}}{} for Python which is both compatible with
Tensorflow \cite{tensorflow2015-whitepaper} and PyTorch
\cite{paszke2017automatic}, introducing and implementing efficient and scalable algorithms for the analysis of loss surfaces during the training of deep neural networks.
\textit{GradVis} enables the user to
visualize the trajectory of their model as well as the loss landscape and the eigenvalues of its Hessian.
We propose efficient mathematical formulations of the key operators and a novel parallelization scheme that allows the investigation of the optimization process using practically relevant neural networks. First experiments show
how  the combination of visualizing the loss
landscape and plotting the eigenvalue density spectrum leads to further insights
into DNN training. In detail, our contributions are:
\begin{itemize}
	\item Implementing efficient mathematical formulations for the computation of the Eigenvalue spectrum of the Hessian, reducing the complexity.
	\item Introducing a novel method-parallelization computation of the Eigenvalue spectrum, improving the parallel efficiency from $37\%$ of a data-parallel approach to over $96\%$
	\item We propose a solution to finding good directions in the high dimensional loss landscape for visualization, solving the problem of using a PCA for the approximation of the trajectory, as discussed in \cite{antognini2018pca}.
	\item Using our fast methods, we conducted several experiments that were (to our best knowledge) not possible before:
	\begin{itemize}
		\item Computing the eigenvalue density spectrum at every iteration, crating high resolution videos of the trajectories and second order information while training a neural network.
		\item Plotting a high resolution loss landscape in between two minima and analysing the eigenvalue density spectrum at all intermediate points
	\end{itemize}
\end{itemize}

\newpage
\section{Methods}

\subsection{Filter-wise Normalization for Visualization}
The Toolbox visualizes loss landscapes by projecting it onto two directional vectors in parameter space $\phi_1$ and $\phi_2$. It normalizes them and plots the loss along the plane that is spanned by those
directions:

\begin{equation}
L(\theta_n + \alpha \phi^*_1 + \beta \phi^*_2)
\end{equation} 

where $\theta_n$ corresponds to the converged point in weight space.

Following \cite{DBLP:journals/corr/abs-1712-09913}, the Visualization Toolbox performs filter-wise normalization on the weights of the network to overcome the scale invariance of the parameter space.
Given weights of a convolutional filter $\theta_{ij}$, filter wise-normalization performs the following operation:
\begin{equation}
d_{ij}^* = \frac{d_{ij}}{\norm{d_{ij}}}\norm{\theta_{ij}}
\end{equation}
where $d_{ij}$ are the convolution weights of the directional vector $d$ and
$d_{ij}$ corresponds to the jth filter for the ith convolution inside the
neural network. $\norm{.}$ corresponds to the Frobenius norm. The
directional vector $d$ is either a random Gaussian vector, a custom vector
provided by the user or it is calculated using principal component analysis
(PCA) over all the saved iterations. All batch-norm parameters are set to zero.

PCA is a dimensionality reduction method. It allows finding the direction and
magnitude of maximum data variation to the inverse of the covariance matrix
\cite{borga1997unified}.  The Toolbox can perform a PCA over the training runs
$\left(\theta_1,...,\theta_n\right)$ where $\theta_{i}$ are the networks weights
at a certain iteration. We obtain the first two PCA directions $\phi_1$ and
$\phi_2$.


The pseudocode is described in Algorithm \ref{alg:algo}.

\begin{algorithm}[h!]
	\caption{Calculate the loss landscape for a neural network with loss $L$ for the n weights along the trajectory $(\theta_1,...,\theta_n)$. The resulting 2D landscape has N points along each axis on the grid}
	\label{alg:algo}
	\begin{algorithmic}
		\REQUIRE Weights $(\theta_1,...,\theta_n)$
		\STATE Calculate the PCA vectors
		\STATE $\phi_1$,$\phi_2$ $\leftarrow$ PCA($\theta_1-\theta_n,...,\theta_{n-1}-\theta_n$) 
		\STATE Use filter-wise normalization
		\STATE $\phi^*_1$ $\phi^*_2 \leftarrow$ Normalize($\phi_1,\phi_2$) 
		\STATE Calculate the coefficients of the training path
		\STATE $\alpha_i,\beta_i \leftarrow $ Solve for $\alpha_i\phi^*_1+\beta_i\phi^*_2 = \theta_{i}-\theta_n$ for each i
		\STATE Calculate the loss values of the path points
		\STATE $z_i \leftarrow L(\alpha_i\phi^*_1+\beta_i\phi^*_1+\theta_n)$ 
		\STATE Calculate the loss values for each point on the grid
		\STATE Make grid of N samples going from $min(\alpha_i)$ to $\max(\alpha_i)$ for x and  $min(\beta_i)$ to $\max(\beta_i)$ for y
		\FOR{i from 0 to N}
		\FOR{j from 0 to N}
		\STATE $x=i(max(\alpha)-\min(\alpha))/N$
		\STATE $y=j(max(\beta)-\min(\beta))/N$
		\STATE $z_{i,j} \leftarrow L(x\phi^*_1+y\phi^*_2+\theta_n)$
		\ENDFOR
		\ENDFOR
		\RETURN ($z_{1,1},...,z_{N,N}$)
	\end{algorithmic}
\end{algorithm}

\subsection{Stochastic Lanczos quadrature algorithm} The stochastic Lanczos
quadrature algorithm \cite{lanczos} is a method for the approximation of the eigenvalue density of very
large matrices. 
The eigenvalue density spectrum is given by:
\begin{equation}
\phi(t) = \frac{1}{N}\sum_{i=1}^{N}\delta(t-\lambda_i)
\end{equation}

where $N$ is the number of parameters in the network and $\lambda_i$ is the i-th eigenvalue of the Hessian.

The eigenvalue density spectrum is approximated by a sum of Gaussian functions:

\begin{equation}
\phi_{\sigma}(t) = \frac{1}{N}\sum_{i=1}^{N}f(\lambda_i,t,\sigma^2)
\end{equation}

where 

\begin{equation}
f(\lambda_i,t,\sigma^2) = \frac{1}{\sigma \sqrt{2\pi}}\exp(-\frac{(t-\lambda_i)^2}{2\sigma^2})
\end{equation}

We use the Lanczos algorithm with full reorthogonalization in order to compute
eigenvalues and eigenvectors of the Hessian and to ensure orthogonality between
the different eigenvectors.  Since the Hessian is symmetric we can diagonalize
it. All eigenvalues will be real.
The Lanczos algorithm returns a tridiagonal matrix. This matrix is diagonalized:

\begin{equation}
T = ULU^T
\end{equation}

By setting $\omega_i = (U_{1,i}^2)_{i=1}^m$ and $l_i = (L_{ii})_{i=1}^m$, the resulting eigenvalues and eigenvectors are used to estimate the true eigenvalue density spectrum:

\begin{equation}
\hat{\phi}^{(v_i)}(t)=\sum_{i=1}^{m}\omega_if(l_i,t,\sigma^2)
\end{equation}

\begin{equation}
\hat{\phi}_{\sigma}(t) = \frac{1}{k}\sum_{i=1}^{k}\hat{\phi}^{(v_i)}(t) 
\end{equation}

The pseudocode for the stochastic Lanczos quadrature algorithm is shown in Algorithm \ref{alg:quadrat}.

\begin{algorithm}[h!]
	\caption{Stochstic Lanczos quadrature algorithm}
	\label{alg:quadrat}
	\begin{algorithmic}
		\REQUIRE Number of iterations k, number of eigenvalues m
		\STATE Initialize Guassian vectors $(v_1,...,v_k)$
		\FOR{i from 1 to k}
		\STATE Run Lanczos with reothogonalization with $L(v_i)$
		\STATE Obtain tridiagonal matrix T
		\STATE Diagonalize $T = ULU^T$
		\STATE Set $l_i = (L_{ii})_{i=1}^m$ and $\omega_i = (U_{1,i}^2)_{i=1}^m$
		\STATE Compute $\hat{\phi}^{(v_i)}(t)=\sum_{i=1}^{m}\omega_if(l_i,t,\sigma^2)$
		\ENDFOR
		\STATE Compute average $\hat{\phi}_{\sigma}(t) = \frac{1}{k}\sum_{i=1}^{k}\hat{\phi}^{(v_i)}(t) $
		\RETURN  $\hat{\phi}_{\sigma}(t)$
	\end{algorithmic}
\end{algorithm}

\subsection{Parallelization}
\textbf{Data-Parallel Visualization.}
The visualization method is trivially parallelizable by assigning parts of the evaluation grid to different
workers. After each worker is done computing its values, the
master-worker collects the values of each worker using MPI\_Gatherv. The
peudocode for the parallelized version is shown in Algorithm \ref{alg:algopar}.

\begin{algorithm}[h!]
	\caption{Parallel visualization method. Calculate the loss landscape for a neural network with loss $L$ for the n weights along the trajectory $(\theta_1,...,\theta_n)$. The resulting 2D landscape has N points along each axis on the grid}
	\label{alg:algopar}
	\begin{algorithmic}
		\REQUIRE Weights $(\theta_1,...,\theta_n)$
		\STATE Calculate the PCA vectors
		\STATE $\phi_1$,$\phi_2$ $\leftarrow$ PCA($\theta_1-\theta_n,...,\theta_{n-1}-\theta_n$) 
		\STATE Use filter-wise normalization
		\STATE $\phi^*_1$ $\phi^*_2 \leftarrow$ Normalize($\phi_1,\phi_2$) 
		\STATE Calculate the coefficients of the training path
		\STATE $\alpha_i,\beta_i \leftarrow $ Solve for $\alpha_i\phi^*_1+\beta_i\phi^*_2 = \theta_{i}-\theta_n$ for each i
		\STATE Calculate the loss values of the path points
		\STATE $z_i \leftarrow L(\alpha_i\phi^*_1+\beta_i\phi^*_1+\theta_n)$ 
		\STATE Calculate the loss values for each point on the grid
		\STATE Make grid of N samples going from $min(\alpha_i)$ to $\max(\alpha_i)$ for x and  $min(\beta_i)$ to $\max(\beta_i)$ for y
		\STATE Split grid into m chunks according to number of workers m $X_k$,$Y_k$ for k = 1,...,m
		\STATE Assign each worker part of the grid 
		\FOR{$x$ in $X_k$}
		\FOR{$y$ in $Y_k$}
		\STATE $z_{x,y} \leftarrow L(x\phi^*_1+y\phi^*_2+\theta_n)$
		\ENDFOR
		\ENDFOR
		\STATE Each worker has set of values $Z_k$
		\STATE $Z$ $\leftarrow$ MPI\_Gatherv($Z_k$)
		\RETURN Z = ($z_{1,1},...,z_{N,N}$)
	\end{algorithmic}
\end{algorithm}

\textbf{Data-Parallel Lanczos.}
The Lanczos algorithm only requires the Hessian-vector product instead of the
full Hessian matrix in order to compute eigenvalues.  This allows using the
R-operator to calculate these products effieciently. Using MPI, we distribute
the batches to different workers, where each worker accumulates its own
Hessian-vector products by using the R-operator. After all workers are done
computing their batches the resulting vectors are sent to the master-worker and
summed by using the MPI\_Reduce operation. The pseudocode is shown in Algorithm
\ref{alg:rop}.

\begin{algorithm}[h!]
	\caption{Data parallel Lanczos. Calculation of Hessian-vector products $Hv$ by using the R-operator}
	\label{alg:rop}
	\begin{algorithmic}
		\REQUIRE vector $v$ and neural network model $h(x)$
		\STATE Set batch size B and therefore divide the dataset into p batches
		\FOR{i from 1 to p}
		\STATE Worker j grabs batch and starts computing the hessian vector product $w_i = R_{op}(h(b_i),v)$
		\ENDFOR 
		\STATE The resulting Hessian vector product gets accumulated during this loop for each worker $w_j = \frac{1}{|I_j|}\sum_{r \in I_j}w_r$ where $I_j$ contains the indices of batches that were computed for worker j
		\STATE Send all resulting $w_j$ to the master-worker using $w = MPI\_Reduce(w_j)$
		\RETURN  Hessian vector product over all samples $w$
	\end{algorithmic}
\end{algorithm}

This data parallel approach allowed us to use far more samples in our
eigenvalue spectrum computation. Future research could look into model
parallelism as well, since the R-operator works exactly like a neural network
with forward and backward propagation.

\textbf{Novel Iteration-Parallel Lanczos.}
We introduce another approach to parallelize the Lanczos method, that proved to be much more
scalable. The basic idea is to let each worker compute one iteration of the stochstic
Lanczos quadrature algorithm for different initializations and then
accumulating all the results at the end on the master-worker. This approach can
be seen in Algorithm \ref{alg:quadratpar}.

\begin{algorithm}[h!]
	\caption{Parallel stochstic Lanczos quadrature algorithm with MPI}
	\label{alg:quadratpar}
	\begin{algorithmic}
		\REQUIRE Number of iterations k, number of eigenvalues m
		\STATE Initialize Guassian vectors $(v_1,...,v_k)$ and split this set to $s$ different workers
		\FOR{$v_i$ from the set assigned to each worker}
		\STATE Run Lanczos with reothogonalizationon on worker w with $L(v_i)$
		\STATE Obtain tridiagonal matrix T
		\STATE Diagonalize $T = ULU^T$
		\STATE Set $l_i = (L_{ii})_{i=1}^m$ and $\omega_i = (U_{1,i}^2)_{i=1}^m$
		\ENDFOR
		\STATE Each worker sends its computed $l_i$ from all different initializations to the master-worker using MPI\_Gatherv. The same is done for the $\omega_i$
		\STATE Compute average on the master-worker 
		$\hat{\phi}_{\sigma}(t) = \frac{1}{k}\sum_{i=1}^{k}\sum_{i=1}^{m}\omega_if(l_i,t,\sigma^2)$
		\RETURN  $\hat{\phi}_{\sigma}(t)$
	\end{algorithmic}
\end{algorithm}
\section{Experiments}

\subsection{Timing and Scalability}
As seen in Algorithm \ref{alg:algo}, the toolbox evaluates the network on the
training samples $N^2$ times, where N is the number of points on the grid along
the x- and y-axis. On each grid point, the network evaluates $n_b$ batches and
the evaluation of each batch takes time $T_{inference}$. Hence, the time to evaluate all the grid points is of the order
of $\alpha + N^2T_{inference}n_b$, with some overhead $\alpha$ for the computation of the
PCA components, as well as for performing the filter-wise normalization. But this
overhead is small for large enough $N$. 
For example, performing experiments with a Resnet-32 for a batch size of 256 and 2 batches resulted in $\alpha = 8s$ and $T_{inference} = 0.005s$.
Table \ref{tab:tab1} shows, that the overhead $\alpha$ stays constant while computation time for the loss landscape scales like $N^2$
\begin{table}[h!]
	\centering
	\caption{Timing of loss landscape calculation and overhead for different N}
	\label{tab:tab1}
	\begin{tabular}{ lcccc }
		\multicolumn{5}{c}{Timing in seconds} \\
		& N=2 &N=10&N=50&N=100\\
		\cline{2-5}
		$N^2T_{inference}n_b$   & 0.51    &12.72&   320& 1890\\
		$\alpha$&   8.21  & 8.01   & 8.09&8.11\\
		
	\end{tabular}
	
\end{table}

\begin{figure}[h!]
	\includegraphics[width=\linewidth]{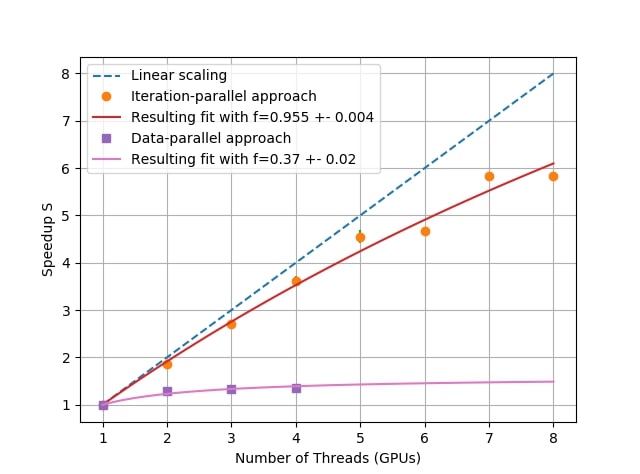}
	\centering
	\caption{Strong scaling plot for the stochastic Lanczos quadrature algorithm with data parallelism and with our proposed method.}
	\label{fig:bothlanscaling}
\end{figure}

\textbf{Scalability.}
The scalability is measured by measuring the speedup given by 
\begin{equation}
S = \frac{T_1}{T_p},
\end{equation}
where $T_1$ is the time for one process and $T_p$ the time for $p$ processes.
In our setting each node contains two Nvidia GTX 1080ti GPUs and they
are assigned sequentially by first filling one node and then populating the
next node with increasing number of ranks. 
The standard deviation on the datapoints is calculated as follows:
\begin{equation}
\sigma_s = \sqrt{(\frac{1}{T_p})^2\sigma_{T_1}^2+(\frac{T_1}{T_p^2})^2\sigma_{T_p}^2}
\end{equation}
We measure strong scaling by keeping the problem size fixed and varying the number of GPUs, computing the parallelizable fraction according to Amdahls law:
\begin{equation}
\label{eq:amdahl}
S = \frac{1}{(1-f)+f/p},
\end{equation}
where $f$ is the parallelizable fraction of our implementation and $p$ refers to the number of GPUs working in parallel on the problem. 
For the stochstic Lanczos quadrature algorithm presented in Algorithm \ref{alg:quadratpar}, the scaling plot is shown in Figure \ref{fig:bothlanscaling}.
One can see that fitting Formula \ref{eq:amdahl} to our data we obtain a parallelizable fraction of $f = 95.5 \pm 0.4 \%$.
Comparing this to the data parallel approach, here we obtain the strong scaling plot in Figure \ref{fig:bothlanscaling}.

Fitting the model yields a parallelizable fraction of $f = 37 \pm 2 \%$, which much is worse than our novel method.

\begin{figure}[h!]
	\includegraphics[width=\linewidth]{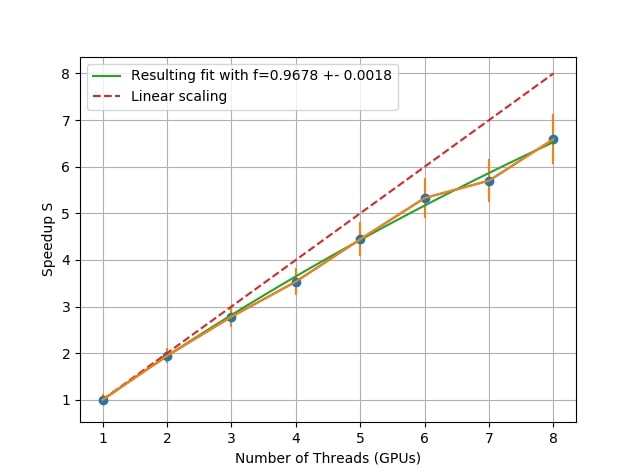}
	\centering
	\caption{Strong scaling plot for the visualization algorithm.}
	\label{fig:visscaling}
\end{figure}

For the visualization presented in Algorithm \ref{alg:algopar}, we obtain Figure \ref{fig:visscaling}.

Here we see that we get a parallelizable fraction of $f=96.78 \pm 0.18 \%$. Due
to the trivially parallelizable nature of the algorithm at hand, the
parallelizable fraction is very high.

\subsection{Finding directions for Visualization}
\begin{figure}[!htbp]
	\begin{minipage}{0.9\linewidth}
		\includegraphics[width=\linewidth]{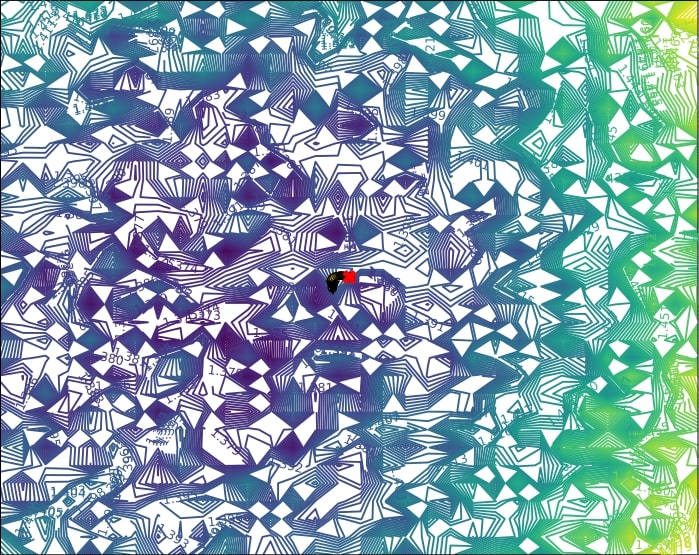}
		\caption{Loss landscape with trajectory along two random directions for LeNet on CIFAR10.}
		\label{fig:rand}
	\end{minipage}
	
	\begin{minipage}{0.9\linewidth}
		\includegraphics[width=\linewidth]{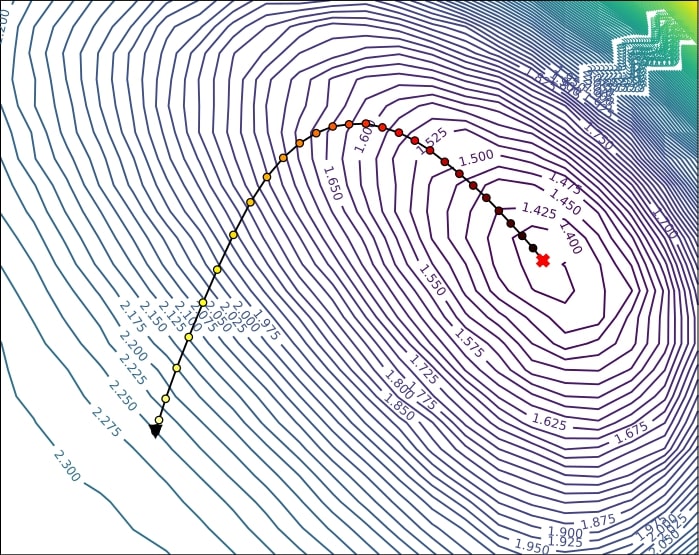}
		\caption{Loss landscape with trajectory of the same training run, visualized along two PCA directions with highest variance for LeNet on CIFAR10, as suggested by \cite{DBLP:journals/corr/abs-1712-09913}. }
		\label{fig:pca}
	\end{minipage}
	
	\begin{minipage}{0.9\linewidth}%
		\includegraphics[width=\linewidth]{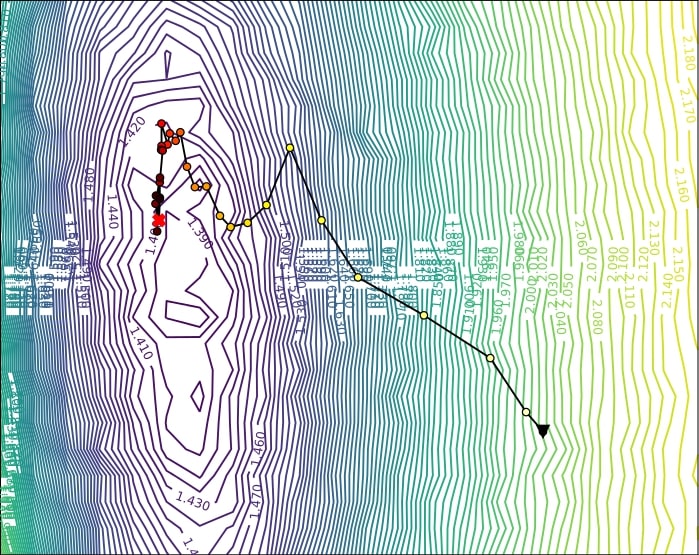}
		\caption{Loss landscape with trajectory of the same training run, visualized along eigenvectors corresponding to the two highest eigenvalues for LeNet on CIFAR10.}
		\label{fig:pospos}
	\end{minipage}
\end{figure}

In order to circumvent the issue of using PCA on the training data (as discussed by \cite{antognini2018pca}), we propose
using different eigenvectors to plot the loss landscape together with the
trajectory. Here we make use of the Lanczos algorithm that we combined with the
R-operator in order to efficiently compute Hessian-vector products. Since the
Lanczos algorithm is an iterative eigenvalue solver we are able to pick certain
eigenvectors we are interested in. Therefore, instead of choosing random
directions or PCA directions we propose using eigenvectors in order to
visualize the training trajectory on the loss landscape.
We compare the different methods to find directions in order to plot the
trajectory. First we choose two randomly initialized vectors with entries drawn
from a normal distribution. The resulting plot is shown in Figure
\ref{fig:rand}.
Next we use PCA to extract the two directions with the highest variance. The resulting visualization is shown in Figure \ref{fig:pca}.
Lastly we use our proposed method of choosing eigenvectors to plot the trajectory. Here we choose the eigenvectors corresponding to the highest two eigenvalues. The resulting plot can be seen in Figure \ref{fig:pospos}.

We can see how the random direction method in Figure \ref{fig:rand} provides
little to no information on what happens during training. The loss landscape is
very noisy and relatively flat, the training path barely moves from its initial
position.

On the contrary, the PCA directions show the trajectory spiraling into the
minimum, which is clearly visible. The path that the trajectory takes is
exactly what one would expect \cite{antognini2018pca}, if a PCA is taken on a high-dimensional random
walk with drift. Therefore even though it seems that the trajectory offers
insight into the training procedure, in reality we will always get the same
trajectory, thereby rendering this method useless.

Our proposed method on the other hand, shows how the trajectory moves along the
gradient and after reaching the valley slowly creeps towards the minimum. This
method also allows picking interesting eigenvalues where our network seems to
"fail" during training, such as negative or zero eigenvalues. This allows
monitoring what the training trajectory was doing in these directions.

\subsection{Loss Landscape with Trajectory and Eigenvalue Density for LeNet}
\label{subsec:losslandsci}
We present the loss landscape together with the trajectory of a LeNet
architecture  that is trained on CIFAR10 \cite{CIFAR}. The neural network is
trained for 1 epoch with 196 iterations with a batch size of 256 and a learning
rate of 0.001, using SGD. Also a momentum of 0.9 is used. After each iteration
the weights of the model are saved onto the hard drive.

\begin{figure*}[!htb]
	\centering
	\begin{subfigure}[t]{0.32\textwidth}
		\includegraphics[width=\linewidth]{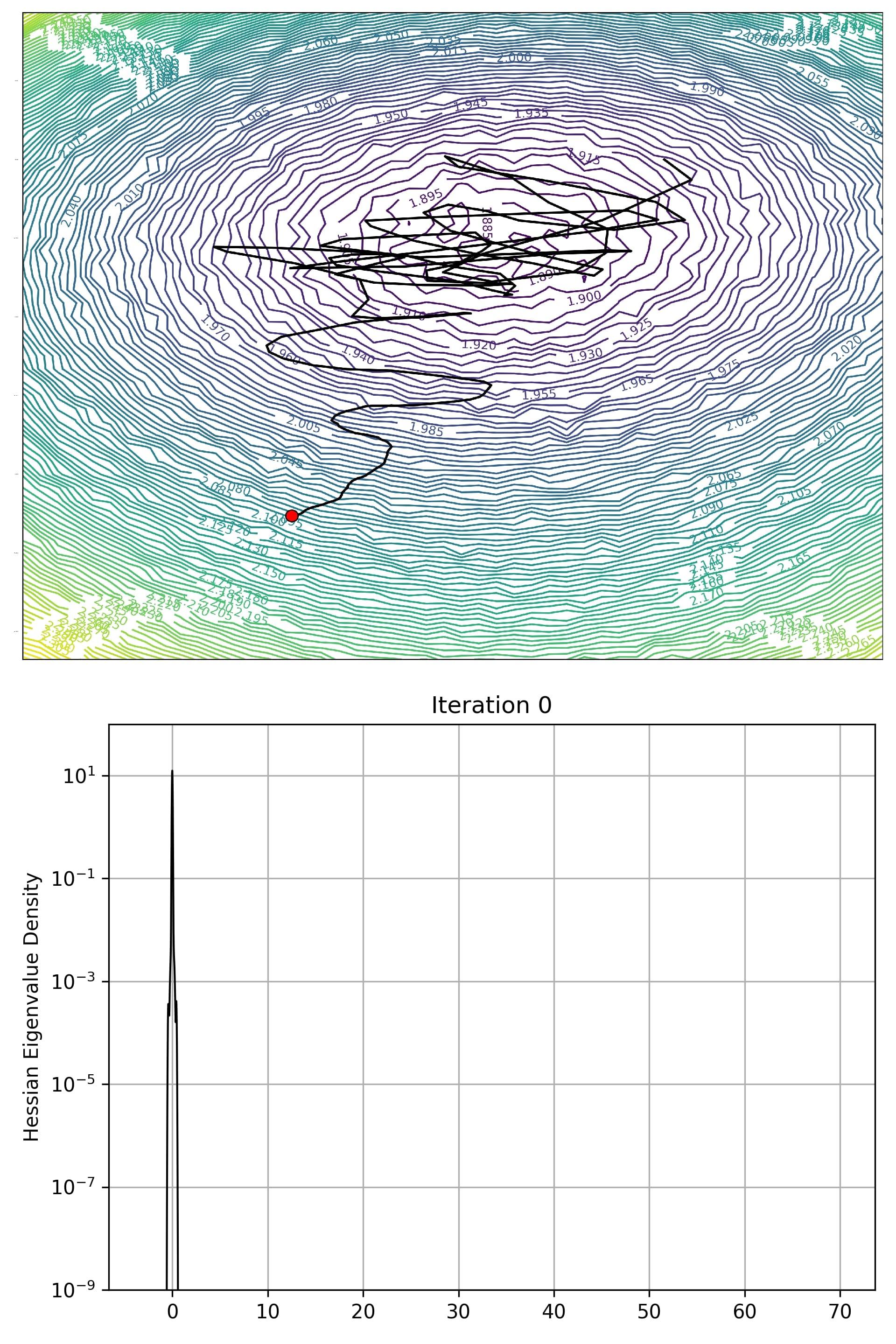}
		\caption{Loss landscape and eigenvalue density plot of LeNet at initialization.}
		\label{fig:iter0}
	\end{subfigure}\hfill
	\begin{subfigure}[t]{0.32\textwidth}
		\includegraphics[width=\linewidth]{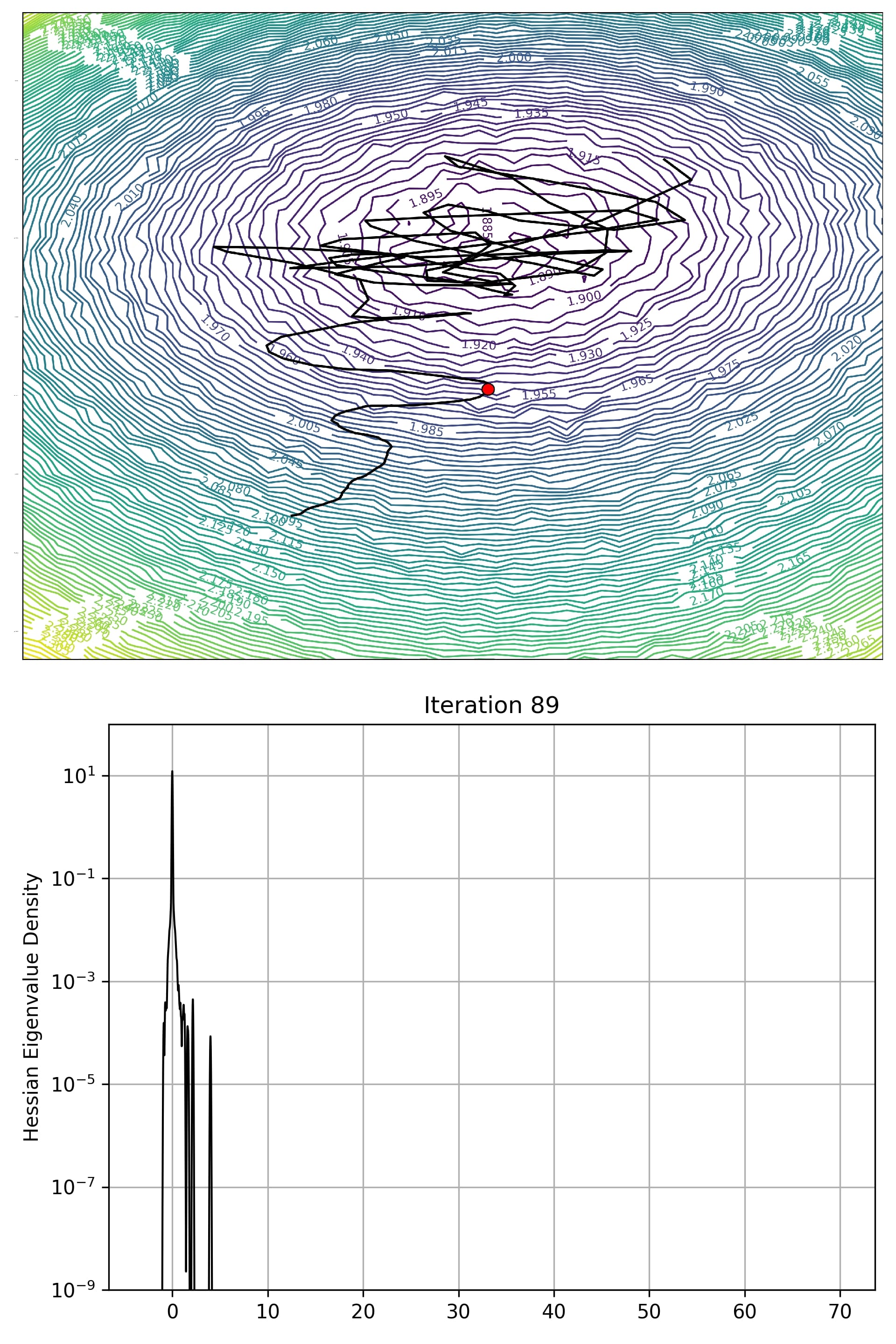}
		\caption{Loss landscape and eigenvalue density plot of LeNet. Same training run at 89 iterations.}
		\label{fig:iter1}
	\end{subfigure}\hfill
	\begin{subfigure}[t]{0.32\textwidth}
		\includegraphics[width=\linewidth]{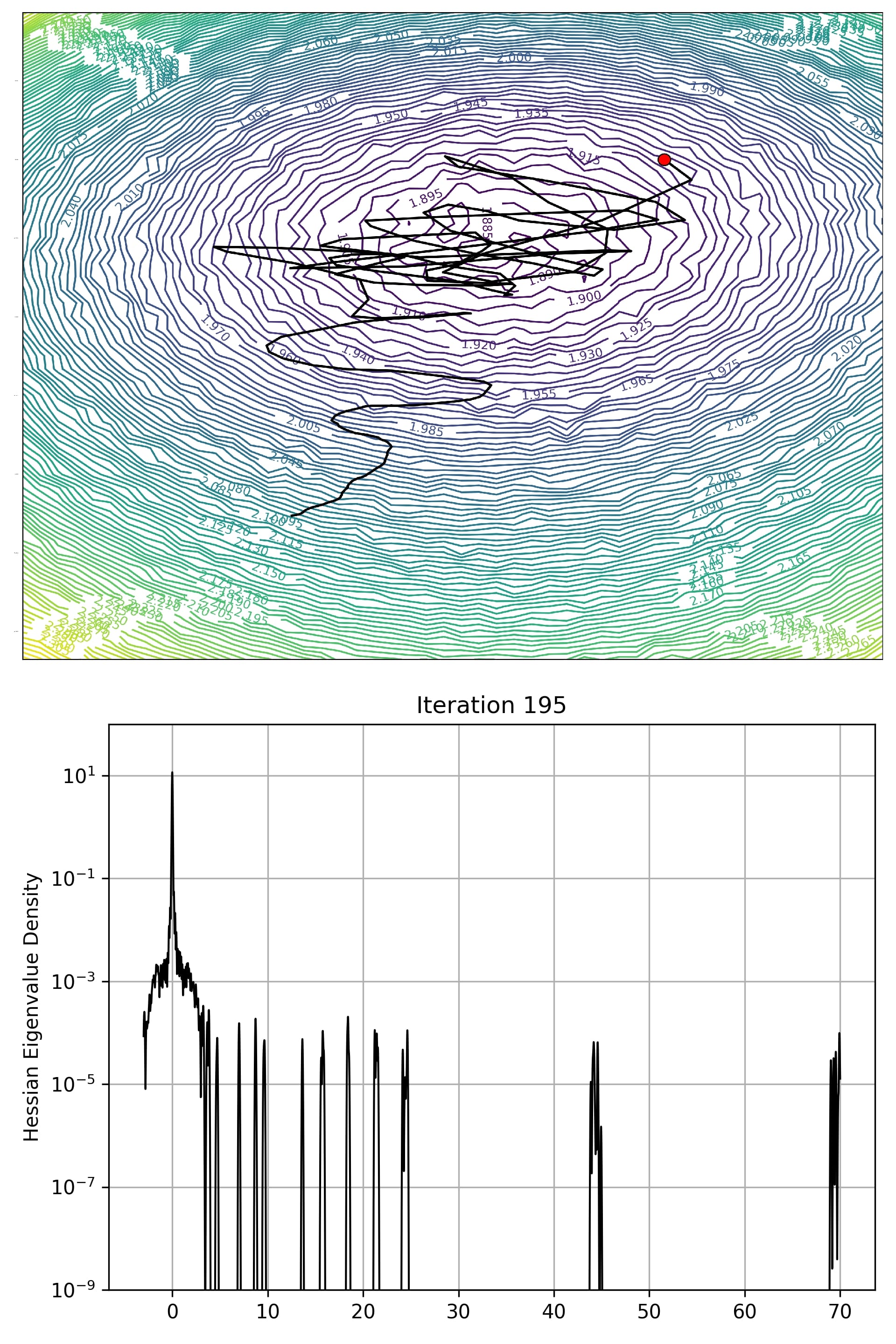}
		\caption{Loss landscape and eigenvalue density plot of LeNet. Same trainig run at 195 iterations.}
		\label{fig:iter2}
	\end{subfigure}
	\caption[LeNet iteration visualization]{Loss landscape and Hessian eigenvalue density for three out of 195 iterations. Here a LeNet architecture is trained in CIFAR10 with using SGD with momentum. A learning rate of 0.01 was used as well as a momentum of 0.9. Both plots were generated using 20 \% of the training samples. The visualization is depicted along two eigenvectors corresponding to the two highest eigenvalues of the Hessian. The plot has a grid size of 50 on each side.}
	\label{fig:iterate_all}
\end{figure*}

For visualization a grid size of N=50 is used with an additional border of 40
\% around the trajectory. The evaluation uses $20\%$ of all the samples
in the training set. The loss landscape is plotted along the two eigenvectors corresponding to the two highest eigenvalues of the Hessian. For the eigenvalue density plots $k=10$ iterations were
used and $m=80$ iterations were performed in the Lanczos algorithm. Also 20\%
of the CIFAR10 samples were used.
The resulting plots are shown in Figure \ref{fig:iterate_all}. These
only show a select number of plots out of the 196 generated \footnote{A video of the full
	trajectory can be viewed on \href{https://youtu.be/0AKSjp-SHlo}{https://youtu.be/0AKSjp-SHlo}}.


At the beginning, during initialization the network finds itself inside a
relatively flat saddle point with all eigenvalues being almost zero. The
network struggles to escape this flat saddle point, it only manages to converge
towards minima in some directions after 89 iterations, as seen in Figure
\ref{fig:iter1}.


One can see in Figure\ref{fig:iter2}, at the end of the first epoch after 196
iterations, the network has converged into minima in some dimensions, while
overall the network still sits on a saddle point. The bulge on eigenvalues
around zero has spread out a little bit more, meaning that in some dimensions
the network sits at a maximum. Looking at the loss landscape, the network
converged into a minimum in the depicted two dimensions. The trajectory oscillates 
around the minimum and it looks like the trajectory is about to escape the 
minimum again, which could be due to the fixed learning rate and the momentum 
which was used in SGD. Also, it seems that the loss landscape is convex in this
area.

\subsection{Interpolation in between two minima}
\begin{figure*}[!htb]
	\centering
	\begin{subfigure}[b]{0.32\textwidth}
		\includegraphics[width=\linewidth]{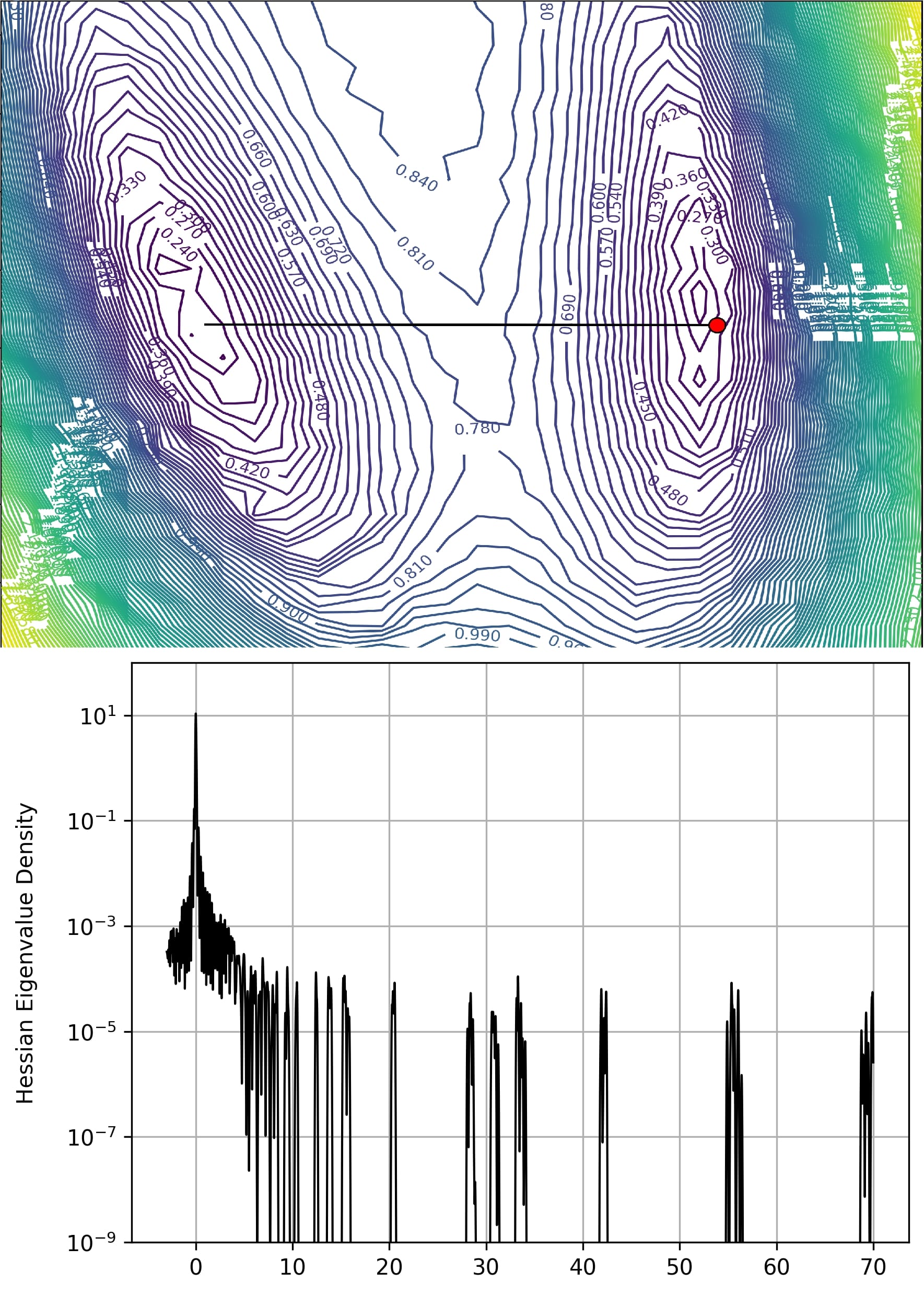}
		\caption{Interpolation between 2 minima with eigenvalue density spectrum in minimum 1}
		\label{fig:interpolmin1}
	\end{subfigure}\hfill
	\begin{subfigure}[b]{0.32\textwidth}
		\includegraphics[width=\linewidth]{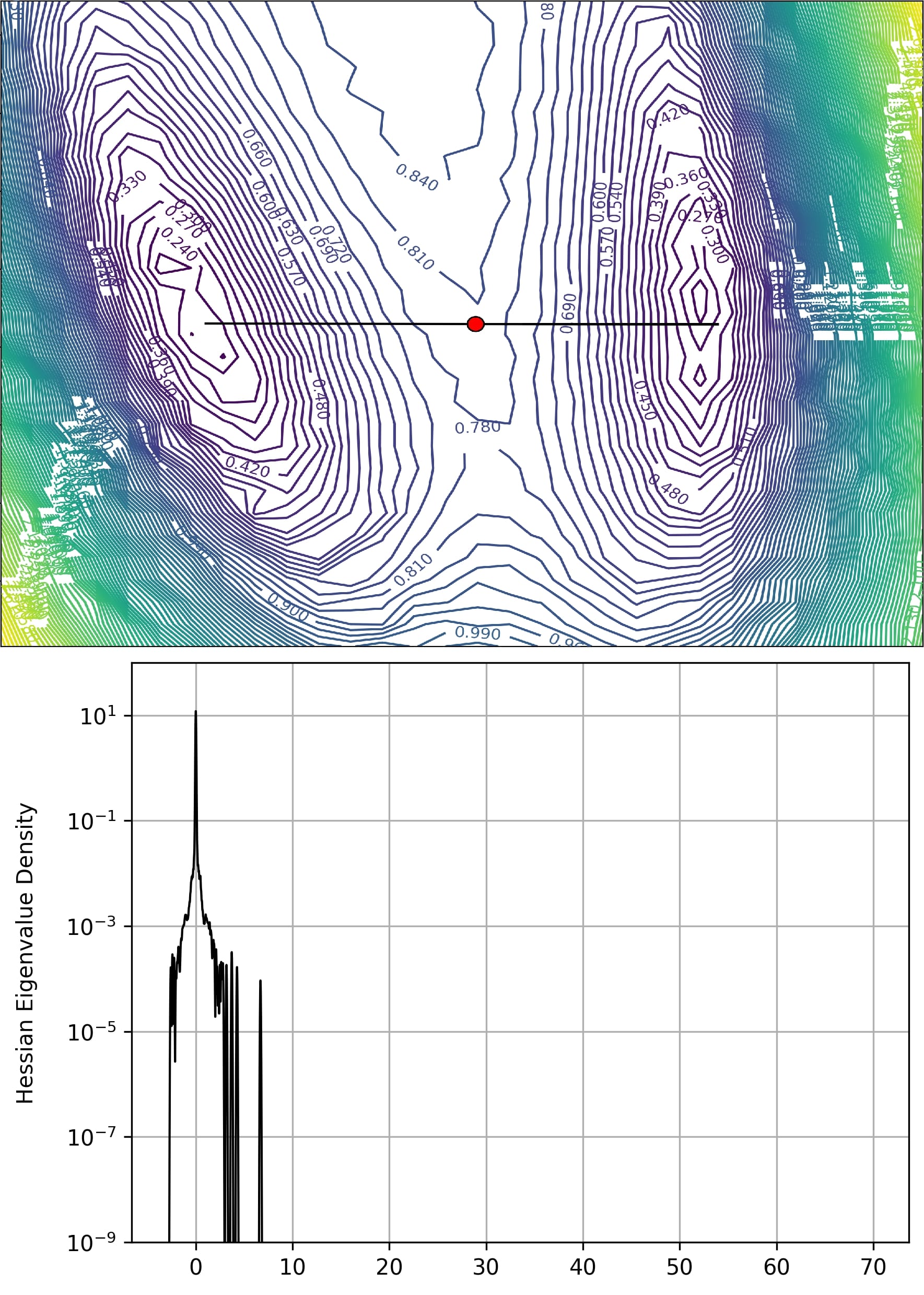}
		\caption{Interpolation between 2 minima with eigenvalue density spectrum in between those}
		\label{fig:interpolmiddle}
	\end{subfigure}\hfill
	\begin{subfigure}[b]{0.32\textwidth}
		\includegraphics[width=\linewidth]{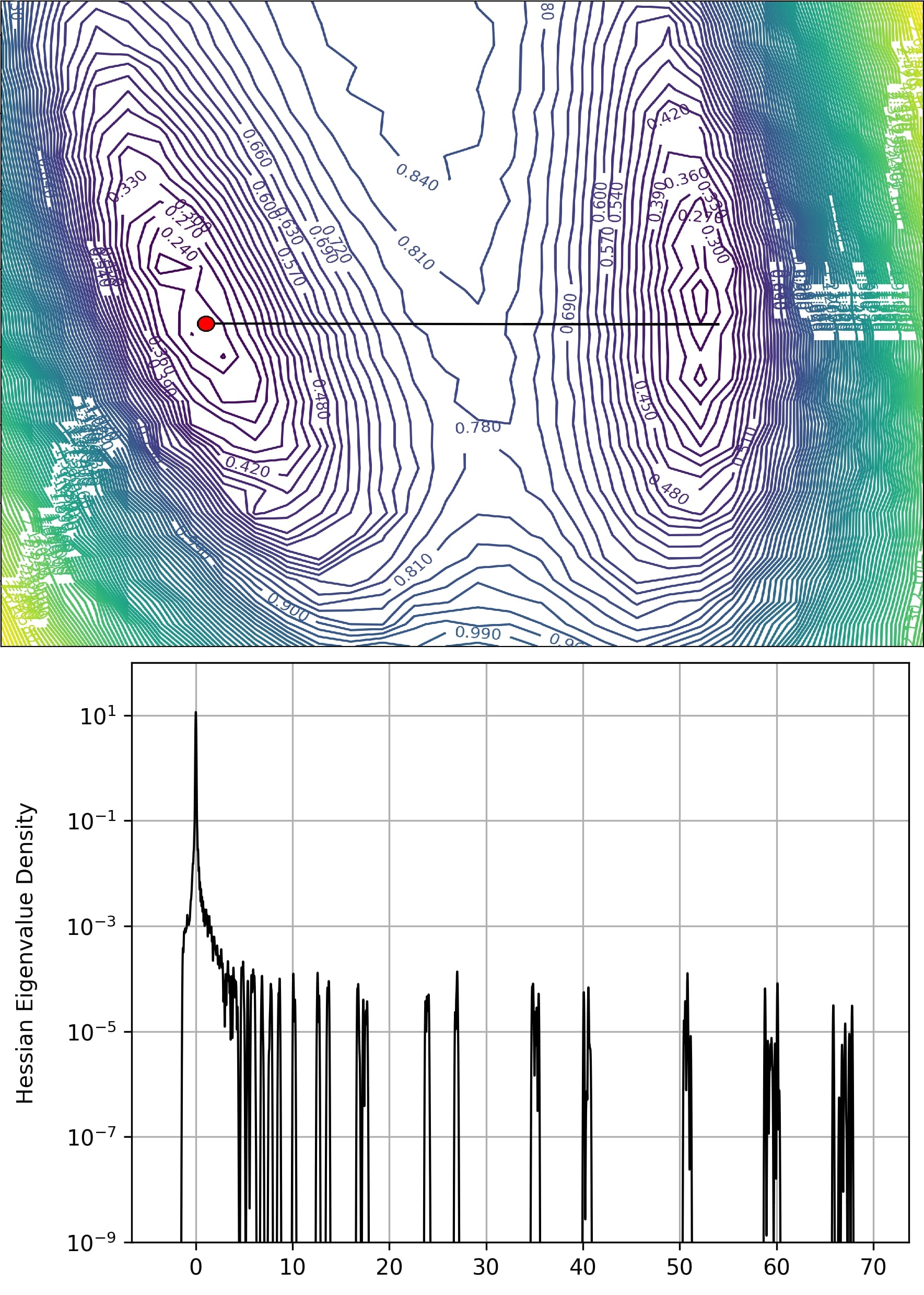}
		\caption{Interpolation between 2 minima with eigenvalue density spectrum in minimum 2}
		\label{fig:interpolmin2}
	\end{subfigure}
	\caption[Interpolation in between two minima]{Visualization of the loss landscape and the Hessian eigenvalue density at different points on the line connecting both minima. A LeNet architecture has been trained with two different initializations to end up in two different minima. For both plots 20 \% of the CIFAR10 training samples were used. The loss landscape plot was crated with a grid size of 50 on each side.}
	\label{fig:interpolall}
\end{figure*}

We train a LeNet network two times with different initializations on 
CIFAR10, using a SGD optimizer with a learning rate of
$lr=0.01$ and a momentum of $m=0.9$. The batch size is $b=256$ and the network
is trained for 10 epochs.
The resulting minima are plotted using the visualization method with a grid
size of $N=50$ and 20\% of all samples. Both minima are connected by a straight
line. We extract 20 points along the line connecting both minima. Using the
stochastic lanczos quadrature with the same parameters as in section
\ref{subsec:losslandsci}, we obtain Figure
\ref{fig:interpolall}\footnote{A video showing all
	points connecting the minima can be viewed on
	\url{https://youtu.be/8UIwPV6yU6I}}.

As show in Figure \ref{fig:interpolall}, the eigenvalues inside both minima indicate, that the network is
sitting on a saddle point, as there are always small negative eigenvalues
present. Also there are a few eigenvalues that are much bigger, indicating a
sharp minimum in the corresponding eigenvector directions. Going toward the
middle of the interpolation those big eigenvalues shift back towards zero. This
shows that in between those minima we are still in a saddle point but there is
no big maximum that separates them, which would be indicated by large negative
eigenvalues.

\section{Discussion}
Looking at the training of a LeNet architecture during its first epoch in
Figure \ref{fig:iterate_all}, one can observe that the network starts
off at a flat saddle point and only gradually falls into minima in some
directions. In most directions the network has zero eigenvalues and some are
also negative. The negative eigenvalues also get more negative during training,
which indicates that the network is located on a maximum in the direction of
the corresponding eigenvector.

Comparing the different methods to plot the loss landscape together with the
trajectory, we see that choosing random directions result in plots that show
little to no information. Here the trajectory barely moves away from its
initial position.  The PCA directions seem to offer a good choice to plot the
trajectory. But PCA actually chooses its directions in such a way that one
always obtains the same figure of the trajectory. Therfore this method is
insufficient in offering any useful information of the training procedure. 
On the other hand, one can see that choosing eigenvectors results in
interesting directions where minima as well as trajectories are present. Here
the trajectories always look different and represent the "true" path taken by
the optimizer.  Looking at the eigenvalues of the interpolation between two
different minima in
Figure \ref{fig:interpolall} one can observe that the
area in between those two minima is relatively flat. Most positive eigenvalues
get pushed toward zero while the negative ones are not changing by a lot
compared to the minima.  Also looking at the scalability of the algorithms, we
see that the visualization method is highly parallelizable, this is to be
expected as we are able to split the grid and each worker can compute its
values independently. For the stochastic Lanczos quadrature algorithm our
algorithm is also highly parallelizable, again this is possible because the
different initializations in the outer loop of the algorithm are independent of
each other. On the other hand, using data parallelism in this algotrithm is
much less parallelizable. One reason could be that after each time the dataset
has been computed, all GPUs have to wait on the algorithm to finish the rest of
the computations for this iteration. Still, if one attempts to scale to
hundreds of GPUs, the best approach would be to perform a mix of both
approaches, as the number of independent iterations in the stochastic Lanczos
quadrature algorithm is on a scale of $10^1$. So using more GPUs than
iterations to compute would make no sense, therefore the remaining ones could
compute samples in a data parallel fashion.

\section{Conclusion}
The presented GradVis tool box now makes it feasible to compute high resolution 
(in terms of optimization space and iterations) visualizations of the loss surfaces, 
gradient trajectories and eigenvalue spectra of the Hessians of full training runs of
practically relevant deep neural networks. Especially our fast implementation and novel
scalable parallelization of the Lanczos algorithm will make GradVis a valuable tool 
towards a better theoretical understanding of SDG based optimizations methods for 
deep learning. First experimental results already show very interesting insights. The next
step for future work will be to compute full scale evaluations for large prominent 
networks on large data sets, like ResNet on ImageNet.

\bibliographystyle{abbrv}
\bibliography{IEEEabrv,bibli}
\end{document}